\documentclass{article}

\usepackage{microtype}
\usepackage{multirow}
\usepackage{graphicx}
\usepackage{subfigure}
\usepackage{dcolumn}
\usepackage[ruled,vlined]{algorithm2e}
\usepackage{amsmath, amssymb, mathtools, bm, amsthm}
\usepackage{booktabs} % for professional tables
\usepackage{todonotes}
\usepackage{hyperref}

\newcommand\blfootnote[1]{%
  \begingroup
  \renewcommand\thefootnote{}\footnote{#1}%
  \addtocounter{footnote}{-1}%
  \endgroup
}

\usepackage{tikz}
\usetikzlibrary{decorations.pathreplacing,calc}
\newcommand{\tikzmark}[1]{\tikz[overlay,remember picture] \node (#1) {};}

\newcommand*{\AddNote}[4]{%
    \begin{tikzpicture}[overlay, remember picture]
        \draw [decoration={brace,amplitude=0.5em},decorate,ultra thick,red]
            ($(#3)!(#1.north)!($(#3)-(0,1)$)$) --  
            ($(#3)!(#2.south)!($(#3)-(0,1)$)$)
                node [align=center, text width=2.5cm, pos=0.5, anchor=west] {#4};
    \end{tikzpicture}
}%

\usepackage{cleveref}
\newcommand{\vs}[1]{\vspace*{-0.2mm}}

\usepackage[accepted]{icml2020}

\usepackage{pifont}% http://ctan.org/pkg/pifont

\newcommand{\highlight}[1]{\colorbox{blue!10}{#1}}

\definecolor{myred}{rgb}{0.75, 0, 0}
\definecolor{myblue}{rgb}{0, 0, 0.75}
\definecolor{mygreen}{rgb}{0, 0.5, 0}

\icmltitlerunning{Learning to Combine Top-Down and Bottom-Up Signals}

\begin{document}

\twocolumn[
\icmltitle{Learning to Combine Top-Down and Bottom-Up Signals\\in Recurrent Neural Networks with Attention over Modules}
\icmlsetsymbol{equal}{*}

\begin{icmlauthorlist}
\icmlauthor{Sarthak Mittal}{mila}
\icmlauthor{Alex Lamb}{udm}
\icmlauthor{Anirudh Goyal}{udm}
\icmlauthor{Vikram Voleti}{mila,udm}
\icmlauthor{Murray Shanahan}{imp}
\icmlauthor{Guillaume Lajoie}{mila,udm}
\icmlauthor{Michael Mozer}{goog}
\icmlauthor{Yoshua Bengio}{mila,udm}
\end{icmlauthorlist}
% \icmlaffiliation{uber}{Uber ATG}
\icmlaffiliation{udm}{Universit\'{e} de Montr\'{e}al}
\icmlaffiliation{mila}{Mila}
\icmlaffiliation{imp}{Imperial College London}
\icmlaffiliation{goog}{Google Research, Brain Team}

\icmlcorrespondingauthor{Anirudh Goyal, Sarthak Mittal}{anirudhgoyal9119@gmail.com, sarthmit@gmail.com}

\icmlkeywords{Deep learning, modular architectures, attention}

\vskip 0.3in
]



\printAffiliationsAndNotice{}  % leave blank if no need to mention equal contribution

\begin{abstract}
% Modern sequential modelling generally relies on Recurrent Neural Network based architectures that construct a memory to capture all the past information. Typical hierarchical models thus end up relying on hierarchies of memory constructed where at each hierarchy the memory is represented by a single vector. Recent work demonstrates that such form of memory is often not able to model complex relational tasks and lacks generalization capabilities. In light of these shortcomings, we propose a hierarchical modular structure in the memory and utilize key-value based attention to update the memory of each module. This inherent modular structure in the memory, we believe, would be able to better capture the nuances in the data and the relations between them. We test our model on a number of sequential tasks and demonstrate that indeed, this hierarchical modular structure in the memory leads to better generalization capabilities and allow the model to perform well on tasks requiring complex reasoning.

Robust perception relies on both bottom-up and top-down signals.
Bottom-up signals consist of what's directly observed through sensation.
Top-down signals consist of beliefs and expectations
based on past experience and short-term memory, 
eg. how the phrase `peanut butter and~...' will be completed.
The optimal combination of bottom-up and top-down information remains an open question,
but the manner of combination must be dynamic and both
context and task dependent. To effectively utilize the wealth of potential top-down information available, and to prevent the cacophony of intermixed signals in a bidirectional architecture, mechanisms are needed to restrict information flow.
We explore deep recurrent neural net architectures in which bottom-up and top-down signals are dynamically combined using attention.  
Modularity of the architecture further restricts the sharing and communication of information. Together, attention and modularity direct information flow, which leads to reliable performance improvements in perceptual and language tasks, and also improves robustness to distractions and noisy data.  We demonstrate on a variety of benchmarks in language modeling, sequential image classification, video prediction and reinforcement learning that the \emph{bidirectional} information flow improves results over strong baselines.
\vspace{-3mm}

%In many applications, state-of-the-art sequential modelling relies on Recurrent Neural Network endowed with self-attention mechanisms to extract information from temporal data streams. 
%
%This process requires a form of non-parametric memory for self-attention to recall past states. This memory is generally constructed sequentially where each entry in a list represents a past state, stored as a vector. 
%
%Such form of memory is often not able to model complex relational tasks and lacks generalization capabilities.
%
%Recent work demonstrates that that instead, a modular network structure that relies on key-value based attention between modules, performs much better. However, such a memory modularity can be inefficient to maintain, and scales poorly.
%
%In light of these shortcomings, we propose a hierarchical modular structure where attention, and thus memory, are subject to connectivity inductive biases. 
%to update memory of each module. 
%
%The rationale is that memory hierarchy is able to better capture the nuances and relations in sequential data thanks to an added robustness in information flow. 
%
%%We test our model on a number of sequential tasks and demonstrate that indeed, this hierarchical modular memory structure leads to better generalization capabilities and allow the model to perform well on tasks requiring complex reasoning.
\end{abstract}

\vs{3}
\section{Introduction}
\vs{1}

%Deep nets have become highly successful, widely used.  

%However sometimes fail to properly handle noisy or unexpected inputs --> adversarial examples, the Uber car failure, failure to consider overall context.  

%Top-down modulation is important, but when and how it happens needs to depend on context.  

%Our goal is to study the extent to which we can use attention to dynamically control when and how top-down and bottom-up signals are combined.  

%Intro
%Related Work: RIMs, Hierarchical RNN, multi-layer LSTM, Deep Boltzmann Machines.  
%Model: H-RIM.  
%Make a figure
%Experiments
%Conclusion

\begin{figure*}[]
\vspace{0.2cm}
\begin{minipage}{1.0\textwidth}
    \centering
    \label{fig:model-diagram}
    \includegraphics[width=1.0\linewidth]{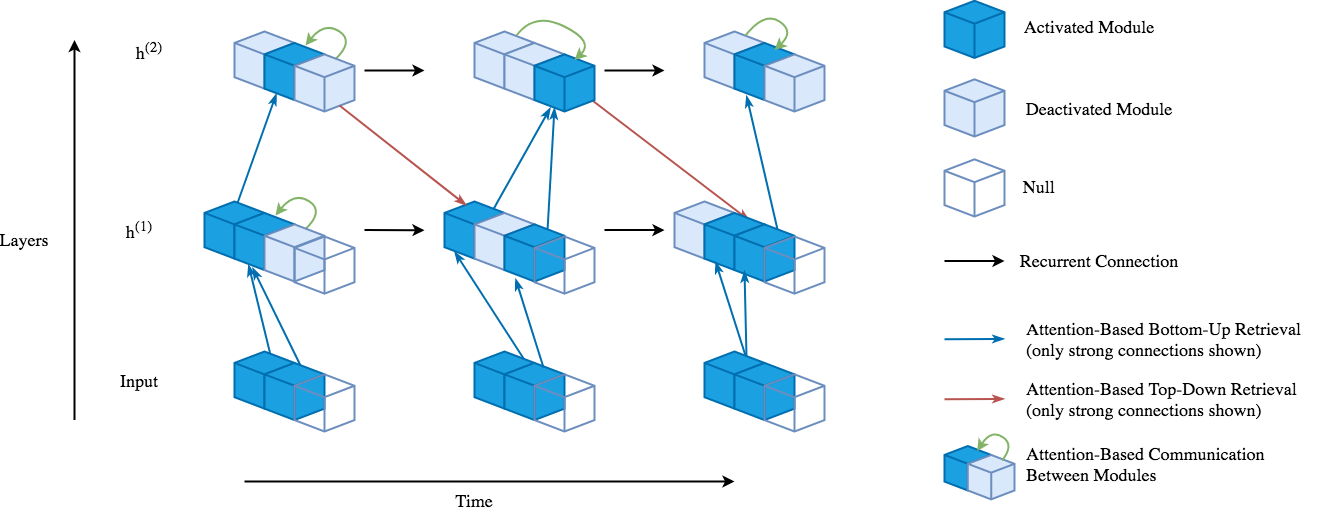}
    \caption{Overall layout of the proposed Bidirectional Recurrent Independent Mechanisms (BRIMs) model.  Information is
    passed forward in time using recurrent connections and information passed between layers using attention.  Modules attend to both the lower layer on the current time step as well as the higher layer on the previous time step, along with \emph{null}, which refers to a vector of zeros.  }
\end{minipage}
\end{figure*}

Deep learning emerged from the goal of learning representational hierarchies,
with the higher levels corresponding to
abstract features used for decision-making~\citep{Hinton06,Bengio-nips-2006-small,salakhutdinov2009deep}.  
Hierarchical models are often taken to imply that computation proceeds in a 
feedforward or \emph{bottom up} fashion, i.e., stage-wise information 
processing in which low-level (sensory) representations construct or modulate
high level (conceptual)  representations. However, they could equally well support the flow of information in a 
feedback or \emph{top down} fashion, i.e., information processing in which
high-level representations modulate lower-level representations. 
 Neuroscientists have noted that
reciprocity of connectivity among anatomically distinct areas 
of neocortex is common \citep{FellemanVanEssen1991,Rockland2015},
causing early visual areas to be modulated by later stages of processing~\citep{Bastos:2015eu}. The same is true for other sensory modalities as well~\citep{Manita:2015bp}.
Neuroimaging research has found evidence for distinct bidirectional activity flows with functional consequences
\citep{Dijkstra2017,Nielsen1999}, and neurophysiological and neuroanatomical studies indicate the important
role of top-down information for guiding processing resources to behaviorally relevant events \citep{baluch2011mechanisms,gilbert2013topdown}. 
The  Global Workspace theory~\citep{baars1997theatre,Dehaene-et-al-2017} posits that
top-down signals are necessary to broadcast information throughout neocortex, which
enables conscious states and verbal behavior.
%also suggests that top-down signals are necessary to broadcoast the information
%selected to pass through the bottleneck of consciousness and reportable
%short-term memory.
Nonetheless, the neural mechanisms of interaction between bottom-up and
top-down pathways are still poorly understood. One goal of our research
is to conduct machine-learning experiments to explore the value of
top-down mechanisms for learning, achieving robustness to distributional shifts,
and guiding the development of novel and improved neural architectures.

In the cognitive science community, the relative contributions of bottom-up and 
top-down signals  have been an ongoing subject of debate for over 40 years
\citep{Kinchla1979,Rauss2013}.
% The computational role of top-down processing was recognized early in the history
% of neural network models in cognitive science. 
The \citet{Mcclelland1981}
model of printed-word reading consisted of a hierarchy of
detectors, from letter fragments to letters to words, with bidirectional 
connectivity. The top-down connectivity 
imposed orthographic constraints of the vocabulary and helped to explain human
proficiency in reading. The model also accounted for puzzling behavioral 
phenomena, such as the fact that people are better at discriminating the letters \textsc{e} 
and \textsc{o} when a letter appears embedded in a word---such as \textsc{r\underline{e}ad}---than
when the same letter is presented in isolation.
In the model, partial activation from 
the representation of the word \textsc{read} provides top-down support for
the activation of the letter \textsc{e} in the context of the word, but not the \textsc{e} in isolation.
%\color{red}Insert text: just as TD helps one letter to cohere with others, TD is thought to achieve the global %coherence of representations necessary to obtain conscious, globally accessible states [citation]\color{black}
If top-down pathways provide expectations, they can
be considered as priors on subsequent inputs, and uncertainty
plays a key role in how information is combined from bottom-up
and top-down sources \cite{Weiss2002,Kersten:2004hd}. For instance, if one enters a familiar but
dark room, one's expectations will guide perception and behavior, 
whereas expectations are less relevant in the light of day.
Even without prior expectations, top-down pathways can leverage 
simultaneous contextual information to resolve local uncertainty,
e.g., the interpretation of the middle letters of these two words:
\includegraphics[]{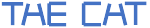}.
Thus, top-down processing leverages goals, temporal and concurrent context, and
expectations (conscious or otherwise) 
%and current conscious content (i.e. verbally reportable) 
to steer or focus
the interpretation of sensory signals. Top-down processing helps to 
overcome intrinsically noisy or ambiguous sensory data, whereas 
bottom-up processing allows an agent to be reactive to unexpected or 
surprising stimuli.

\vspace{-4mm}
\paragraph{Information Flow in Deep Neural Networks.}
Models having within-layer recurrence are common (e.g., LSTM layers), but such recurrence involves a particular 
level of representation interacting with itself, as opposed to the strict top-down notion of higher levels acting on lower levels.
In contrast, a layered architecture with feedforward and feedback connections makes a structural distinction between higher and lower
levels of representations, and it requires a combination of information from two distinct sources.

Bidirectional layered models have long existed \cite{dayan1995helmholtz,Larochelle2008,salakhutdinov2009deep}, 
but these models have not received the same intense scrutiny and development as feedforward models.
The goal of this paper is to revisit bidirectional layered models with the aim of raising them to state-of-the-art in performance and more specifically, out of distribution generalization.
Simply taking existing models and incorporating bidirectional connectivity introduces challenges \cite{Iuzzolino2019}. Instead,
we explore the notion that bidirectional information flow must be coupled with additional mechanisms to 
effectively select from potential top-down signals, modulate
bottom-up signals and prevent a cacophony of intermixed signals.
We propose two specific mechanisms from the deep learning toolkit to dynamically route information flow:  \emph{modularity} and \emph{attention}.
With these two mechanisms, we outperform state-of-the-art feedforward models on challenging vision and language tasks.

\vspace{-4mm}
\paragraph{Dynamic Information Flow Using Attention.}
\citet{goyal2019recurrent} offer evidence that modularity, coupled with a flexible means of communication between modules, can 
%greatly improve performance and, crucially, drastically 
dramatically improve generalization performance.
%of the learned model. 
They describe a recurrent independent mechanisms
(RIMs) architecture that
dynamically controls information flow in a modular RNN. The recurrently connected modules compete
for external input and communicate sets of objects 
%(associatedwith a key and value) 
\emph{sparingly} via differentiable attention
mechanisms driven by the match between keys (from a source module)
and queries (from a destination module). Computation is sparse and modular, meaning that only a subset of the neural modules are active at any time. Composition of computations is dynamic
% and \emph{plug-and-play} 
rather than static, in the sense that 
the attention mechanism uses context to select the subset of modules that are
%which sequence or sub-graph of modules are 
activated and what is communicated to which module, with the input of one module dynamically selected, so the  values associated with selected objects are not always coming from the same modules or parts of the input.
The motivating intuition behind this increased flexibility is that if the relationship between modules changes between training and evaluation, a model which keeps the information in these modules sufficiently separate should still be able to correctly recombine information, even though the overall data distribution differs from what was seen during training.

%as shown in Fig.~\ref{fig:rims_figure}.

%\subsection{Combining Top-Down and Bottom-Up Information with Attention over Modules}
%
\vspace{-4mm}
\paragraph{Combining Top-down and Bottom-up Information with Attention over Modules.}
We now present our central proposal, which is that top-down and bottom-up signals should be combined explicitly and selectively by using attention.  On a given time step, only a fraction of the high-level concepts being detected are likely to be relevant for a particular stimulus.  For example, if a person is trying to classify an object in a dark room (which they are familiar with), the most relevant top-down signal is the person's prior knowledge of what that object looks like, as opposed to other knowledge about the room.  This motivates breaking up the hidden state into multiple modules, such that the top-down and bottom-up interactions can be appropriately focused.  
Thus we consider the use of modules at multiple levels of abstraction to be a key ingredient for our method.  We elect to build upon the Recurrent Independent Mechanisms (RIMs) framework for the modular building blocks, as their end-to-end training demonstrated successful specialization between modules \cite{goyal2019recurrent}. However, while network modularity revealed considerable computational advantages, the modules in the RIMs architecture are fully interconnected and thus there is no notion of layered or hierarchical structure.

%Dynamic control of information flow is fundamental to how brains and computers handle complex problems, whether it be for number crunching in a computer or for sensory processing, motor control, or cognitive processing in the brain. 

%Placing these modules into a multi-layer structure and using attention to communicate between different layers.  This specialization between higher and lower layers may enable effective learning with a greater number of modules.  %To do this we introduce top-down and bottom-up feedback mechanisms should allow this structure to provide better feedback, especially when the number of modules is large.  

To address this issue, we propose a new architecture, which we call \textit{Bidirectional Recurrent Independent Mechanisms (BRIMs)}. This new approach endows the attention-based, modular structure with a hierarchy of layers composed of competing modules, yielding organized contextual computations during training. The rationale is to provide an architectural backbone such that each layer of the hierarchy can send information in both a bottom-up direction as well as in a top-down direction.  We find that this  inductive bias, combined with the modularity mechanisms developed in~\citet{goyal2019recurrent}, provides considerable and further advantages for tasks where there is a change of input distribution from training to testing.

%Our newly proposed architecture, which we call \textbf{``Bidirectional Recurrent Independent Mechanisms'' (BI-RIMs)}, is loosely inspired by the functioning of cognitive processes in the brain, where is is known that higher order areas contain abstract knowledge representations such as concepts, mental models, and schemata, while the bottom of the hierarchy contains concrete, and specific knowledge representations such as visual features, lexicons, and propositions~\re{[CITE]}.  Additionally it is known that different layers of the cortex specialize in either top-down or bottom-up feedback and have distinct structures.  

%We find that this novel inductive bias, combined with the modularity mechanisms developed in~\citet{goyal2019recurrent}, provides considerable advantages for tasks where there is a change of input distribution from training to testing. 
%
%The contribution of our work is as follows:

%\begin{itemize}
%    \item The proposed model presents a novel mechanism that enables higher level layers to selectively query information from lower levels via key-value based attention.
%    \item Modules in higher levels can influence module activation in subsequent time steps and at lower levels, thus enabling contextual information from past states to influence future lower-level processing and computations.
%    \item We demonstrate with numerical experiments how this  approach enables better out-of-distribution generalization performance on several benchmark tasks.
%\end{itemize}

%\vs{2}
\section{Preliminaries}
%\vs{1}

%\begin{figure*}[!htp]
%    \centering
%    \includegraphics[width=\linewidth]{figures/RIMs_base.png}
%    \caption{The Recurrent Independent Mechanism (RIMs) %\citep{goyal2019recurrent} model which we build upon, which lacks explicit top-down connections and has only a single layer of modules. }
%    \label{fig:rims_figure}
%\end{figure*}

%\subsection{Recurrent Neural Networks}
%\label{rnn}
%Sequential modelling of dynamics using RNNs relies on having a memory or hidden state that can provide information about the whole past. Most of the standard recurrent networks thus rely on a vector of fixed dimension $\textbf{h}_t$ to represent the state of the network at time $t$. A general RNN can thus be concisely represented as:
%\begin{align}
%    \textbf{E}_t &= E(\textbf{I}_t) \\
%    \textbf{h}_t &= F(\textbf{E}_t, \textbf{h}_{t-1}) \\
%    \textbf{O}_t &= D(\textbf{h}_t)
%\end{align}
%where $\textbf{I}_t$ denotes the input, $\textbf{O}_t$ the predicted output and $\textbf{h}_t$ the hidden state for any given time $t$. $D$ here represents the decoder that maps hidden states to predictions while $E$ represents the encoder that maps a single input on a step to a representation. $F$ refers to the update dynamics of the hidden state, for eg. an LSTM \citep{hochreiter1997long} or GRU \citep{chung2015recurrent} styled update.

\paragraph{Multi-layer Stacked Recurrent Networks.}
The most common multi-layer RNN architecture is bottom-up and feed-forward, in the sense that higher layers are supplied with the states of the lower layers as inputs. An $L$-layer deep RNN is then concisely summarized as:
%\vs{1.5}
\begin{align}
    \textbf{y}_t &= D(\textbf{h}^L_t) \\
    \textbf{h}^l_t &= F^{l}(\textbf{h}^{l-1}_t, \textbf{h}^l_{t-1}) \label{rnn_hier}  \\
    \textbf{h}^0_t &= E(\textbf{x}_t)
%\vs{1}
\end{align}
with $l = 0,1,...,L$. For a given time $t$, $\textbf{y}_t$ denotes the model prediction, $\textbf{x}_t$ the input and $\textbf{h}^l_t$ the hidden state of the model at layer $l$. $D$ and $E$ denote the Decoder and Encoder for the model. $F^l$ represents the recurrent dynamics at the hierarchy level $l$ (e.g., an LSTM or GRU).

% Typically, RNN-based dynamics models are expressed as functions of the form:

% \begin{equation}
% \vh_{t + 1} = F(\ve_t, \vh_{t}) \qquad \ve_t = E(\vo_t)    
% \end{equation}

% where $\vo_t$ is the observation at time $t$, and $\vh_{t + 1}$ is the hidden state. $F$ can be thought of as parameterized by a neural network such as LSTMs \citep{hochreiter1997long} or  GRUs \citep{chung2015recurrent}. 

%the hidden state $\vh$ conditioned on the observation $\vO$, %whereas $D$ is a \emph{decoder} that maps the hidden state to observations. It is apparent that the evolution function of the modelled dynamical system (as defined in Equation~\ref{eq:dynamic_system_def}) can be obtained by rolling out the forward-evolution function in time. 
% While this class of models are known to be universal approximators of (open) dynamical systems \citep{schafer2006recurrent}, other options exist, e.g. bi-directional RNNs \citep{schuster1997bidirectional} and Transformers \citep{vaswani2017attention}.

%\vs{1}
\paragraph{Key-Value Attention.} 
Key-value Attention (also sometimes called Scaled Dot Product attention), defines the backbone of updates to the hidden states in the proposed model. This form of attention is widely used in self-attention models and performs well on a wide array of tasks \citep{vaswani2017attention, santoro2018relational}. Given a set of queries $\textbf{Q}$, keys $\textbf{K}$ ($d$-dimensional) and values $\textbf{V}$, an attention score $\textbf{A}_S$ and an attention modulated result $\textbf{A}_R$ are computed as
%\vspace{-1mm}
\begin{align}
    \mathrm{\textbf{A}_S} &= \mathrm{Softmax} \left (\frac{\textbf{QK}^T}{\sqrt{d}} \right) \label{att_score}\\
    \mathrm{\textbf{A}_R} &= \mathrm{\textbf{A}_S}\; \textbf{V} \label{att_out}
\end{align}
%\vspace{-10mm}
%Without loss of generality, one may assume that the dynamical system F  can be \emph{factorized} in to constituent systems $(\phi_1, \phi_2, ..., \phi_M)$, such that the interaction between all pairs of sub-systems $(\phi_i, \phi_j)$ satisfy some criterion. Now, the strength of this assumed criterion lies on a spectrum. On the one end of the spectrum is the case where such a criterion is non-existent, i.e. no such factorization is assumed and full generality is restored; this is the modeling assumption made when using 
\paragraph{Recurrent Independent Mechanisms.} RIMs \citep{goyal2019recurrent} consist of a single layered recurrent structure where 
the hidden state $\textbf{h}_t$ is decomposed into $n$ modules, $\textbf{h}_{t,k}$ for $k = 1, ... n$.
%(\hyperref[fig:rims_figure]{Figure~\ref{fig:rims_figure}})
It also has the property that on a given time step, only a subset of modules is activated. 
In RIMs, the updates for the hidden state follow a three-step process. 
First, a subset of modules is selectively activated based on their determination of the relevance of their input. 
Second, the activated modules independently process the information made available to them.
Third, the active modules gather contextual information from all the other modules and consolidate this information in their hidden state.

\paragraph{Selective Activation.} Each module creates queries \mbox{$\bar{\textbf{Q}}={Q}_{inp}\,(\textbf{h}_{t-1})$} ($n \times d$ matrix) which are then combined with the keys $\bar{\textbf{K}}={K}_{inp}\,(\textbf{\o},\textbf{x}_t)$  and values $\bar{\textbf{V}}={V}_{inp}\,(\textbf{\o}, \textbf{x}_t)$ obtained from the input $\textbf{x}_t$ and zero vectors $\textbf{\o}$ to get both the attention score $\bar{\textbf{A}}_S$ and attention modulated input $\bar{\textbf{A}}_R$ as per equations \eqref{att_score} and \eqref{att_out}. Based on this attention score, a fixed number of modules $m$ are activated for which the input information is most relevant (where the null module, which provides no additional information, has low attention score). We refer to this activated set per time-step as $\mathcal{S}_t$.  

\paragraph{Independent Dynamics.} Given the attention modulated input obtained above, each activated module then undergoes an update in its hidden state:
\begin{align}
\bar{\textbf{h}}_{t,k} &= \begin{cases}F_k\, ( \bar{\textbf{A}}_{R_k}\,,\,\textbf{h}_{t-1,k}) \quad & k \in \mathcal{S}_t \\
\textbf{h}_{t-1,k} \quad & k \notin \mathcal{S}_t
\end{cases}
\end{align}
% \begin{align}
% \bar{\textbf{h}}_{t,k} &= F_k\, [ \textbf{A}_R\;(\bar{\textbf{Q}}, \bar{\textbf{K}}, \bar{\textbf{V}})\,,\,\textbf{h}_{t-1,k}] \quad & k \in \mathcal{S}_t \\
% \bar{\textbf{h}}_{t,k} &=\textbf{h}_{t-1,k} \quad & k \notin \mathcal{S}_t
% \end{align}
$F_k$ here stands for any update procedure, e.g. GRU or LSTM. 
% State transformation occurs only for active modules.

\paragraph{Communication.} After an independent update step, each module then consolidates information from all the other modules. RIMs again utilize the attention mechanism to perform this consolidation. Active modules create queries $\hat{\textbf{Q}}=Q_{com}(\bar{\textbf{h}}_{t})$ which act with the keys $\hat{\textbf{K}}={K}_{com}(\bar{\textbf{h}}_{t})$ and values $\hat{\textbf{V}}={V}_{com}(\bar{\textbf{h}}_{t})$ generated by all modules and the result of attention $\hat{\textbf{A}}_R$ is added to the state for that time step:\blfootnote{Code is available at \href{https://github.com/sarthmit/BRIMs}{https://github.com/sarthmit/BRIMs}}

\begin{align}
    \textbf{h}_{t,k} &= \begin{cases}\bar{\textbf{h}}_{t,k} + \hat{\textbf{A}}_{R_k} \quad & k \in \mathcal{S}_t \\
    \bar{\textbf{h}}_{t,k} \quad & k \notin \mathcal{S}_t
    \end{cases}
\end{align}

\section{Proposed Method}

Our contribution is to extend RIMs to a multilayered bidirectional architecture, which we refer to as BRIMs.
Each layer is a modified version of the RIMs architecture \citep{goyal2019recurrent}. A concise representation of our architecture is depicted in \hyperref[fig:model-diagram]{Figure~\ref{fig:model-diagram}}. We now describe
the dynamic bottom-up and top-down flow of information in BRIMs.

%\vs{1}
\subsection{Composition of Modules}
%\vs{1}

%Feels more like an introduction sentence.  
%While there have been advances that introduce different mechanisms for the update of recurrent states \citep{el1996hierarchical, koutnik2014clockwork, chung2016hierarchical, neil2016phased, krueger2016zoneout,  campos2017skip, ke2018focused}, we instead choose to adopt the setting used in RIMs \citep{goyal2019recurrent} that considers a specific structure over the hidden state. 

We use the procedure provided in RIMs to decompose the hidden state $\textbf{h}^l_t$ on each layer $l$ and time $t$ into separate modules. Thus, instead of representing the state as just a fixed dimensional vector $\textbf{h}^l_t$, we choose to represent it as $\{((\textbf{h}^l_{t,k})_{k=1}^{n_l}, \mathcal{S}^l_t)\}$ where $n_l$ denotes the number of modules in layer $l$ and $\mathcal{S}^l_t$ is the set of modules that are active at time $t$ in layer $l$. $|\mathcal{S}^l_t| = m_l$, where $m_l$ is a hyperparameter specifying the number of modules  active in layer $l$ at any time. Each layer can potentially have different number of modules active. Typically, setting $m_l$ to be roughly half the value of $n_l$ works well.

%\vs{1}
\subsection{Communication Between Layers}
\label{within_hier}
%\vs{1}
%\todo{Mozer note:  the choice of blue for the bottom-up arrows is unfortuante because it matches the color of the b oxes. it might be nice to  pick a more distinctive color or a darker blue.}

We dynamically establish communication links between multiple layers using key-value attention, in
a way which differs radically from RIMs \citep{goyal2019recurrent}.  While many RNNs build a strictly bottom-up multi-layer dependency using \eqref{rnn_hier}, we instead build multi-layer dependency by considering queries $\bar{\textbf{Q}}={Q}_{lay}\,(\textbf{h}^l_{t-1})$ from modules in layer $l$ and keys $\bar{\textbf{K}}={K}_{lay}\,(\textbf{\o},\textbf{h}^{l-1}_{t}, \textbf{h}^{l+1}_{t-1})$ and values $\bar{\textbf{V}}={V}_{lay}\,(\textbf{\o},\textbf{h}^{l-1}_{t},\textbf{h}^{l+1}_{t-1})$ from all the modules in the lower and higher layers (blue and red arrows respectively; \hyperref[fig:model-diagram]{Figure~\ref{fig:model-diagram}}). From this three-way attention mechanism, we obtain the attention score $\bar{\textbf{A}}^l_S$ and output $\bar{\textbf{A}}^l_R$ (Equations \ref{att_score} and \ref{att_out} respectively).  Note that in the deepest layer, only the lower layer is used and on the first layer, the input's embedded state serves as the lower layer. Also note that the attention receiving information from the higher layer looks at the previous time step, whereas the attention receiving information from lower layer (or input) looks at the current time step. 
\begin{align*}
\small
    \text{Softmax} \left(\begin{bmatrix}
    \mathbf{Q}_{\;module\; 1} \\
    \mathbf{Q}_{\;module\; 2} \\
    \mathbf{Q}_{\;module\; 3} \\
    \end{bmatrix}
    \begin{bmatrix}
    \mathbf{\color{myred}K}_{\;\phi} & \mathbf{\color{myblue}K}_{\;l-1} & \mathbf{\color{mygreen}K}_{\;l+1}
    \end{bmatrix}
    \right)
    \begin{bmatrix}
    \mathbf{\color{myred}V}_{\;\phi} \\
    \mathbf{\color{myblue}V}_{\;l-1} \\
    \mathbf{\color{mygreen}V}_{\;l+1}
    \end{bmatrix} \\
    \begin{matrix}
    {\color{myred}\phi}: \text{Null} &
    {\color{myblue}l-1}: \text{Bottom Up} &
    {\color{mygreen}l+1}: \text{Top Down}
    \end{matrix}
\end{align*}
%\vspace{-8mm}
\subsection{Sparse Activation}
Based on the attention score $\bar{\textbf{A}}^l_S$, the set $S_t^l$ is constructed which comprises modules for which null information is least relevant. Every activated module gets its own separate version of input (as it's dependent on its query, which is a function of the hidden state of the module) which is obtained through the attention output $\bar{\textbf{A}}^l_R$. Concretely, for each activated module, this can be represented as:
%\vs{1}
\begin{align}
     \bar{\textbf{h}}^l_{t,k} &= F_k^l \, ( \bar{\textbf{A}}^l_{R_k}, \textbf{h}^l_{t-1, k}) \quad k \in \mathcal{S}^l_t
%\vs{1}
\end{align}
where $F_k^l$ denotes the recurrent update procedure.
% For each time step, and for a particular layer $l$, the attention values from $\textbf{A}_R$ are used to select which modules are going to be activated. The modules which have the $k_A$ lowest attention on the zero vector $\textbf{\O}$ per time step (on a given layer) get activated.  $k_A$ is a fixed hyperparameter, but which was found to be flexible in \cite{goyal2019recurrent}, and different per layer.  In many cases, setting $k_A$ to be roughly half the value of $k_T$ performs well.  This set of activated RIM indices constitutes the set $\mathcal{S}^l_t$.  

%For any given module, the score \eqref{att_score} from the  attention determines the affinity of that module to query information from adjacent layers as compared to staying dormant (i.e affinity to access null information $\textbf{\O}$ and), this score creates a competition b/w different modules, such that $k_{i}$ modules are activated at each layer. This creates a communication channel between different layers of the hierarchy where each module of higher hierarchy can query information about the modules at the lower hierarchy and using that information, it can decide whether to activate or not.

%\vs{1}
\subsection{Communication Within Layers}
%\vs{1}

We also perform communication between the different modules within each layer (green arrows; \hyperref[fig:model-diagram]{Figure~\ref{fig:model-diagram}}). In order to enable this communication, we again make use of key-value attention. This communication between modules within a layer allows them to share information, albeit in a limited way through the bottleneck of attention. We create queries $\hat{\textbf{Q}}={Q}_{com}\,(\bar{\textbf{h}}^l_{t})$ from active modules and keys $\hat{\textbf{K}}={K}_{com}\,(\bar{\textbf{h}}^l_t)$ and values $\hat{\textbf{V}}={V}_{com}\,(\bar{\textbf{h}}^l_t)$ from all the modules to get the final update to the module state through residual attention $\hat{\textbf{A}}^l_R$ addition as follows:
%\vs{1}
\begin{align}
    \textbf{h}^l_{t,k} &= \begin{cases}\bar{\textbf{h}}^l_{t,k} + \bar{\textbf{A}}^l_{R_k} \quad&   k \in \mathcal{S}^l_t \\
    \textbf{h}^l_{t-1,k} \quad& k \notin \mathcal{S}^l_t
    \end{cases}
\end{align}
\begin{figure*}[]
\begin{minipage}{1.0\textwidth}
    \centering
    \includegraphics[width=1.0\linewidth]{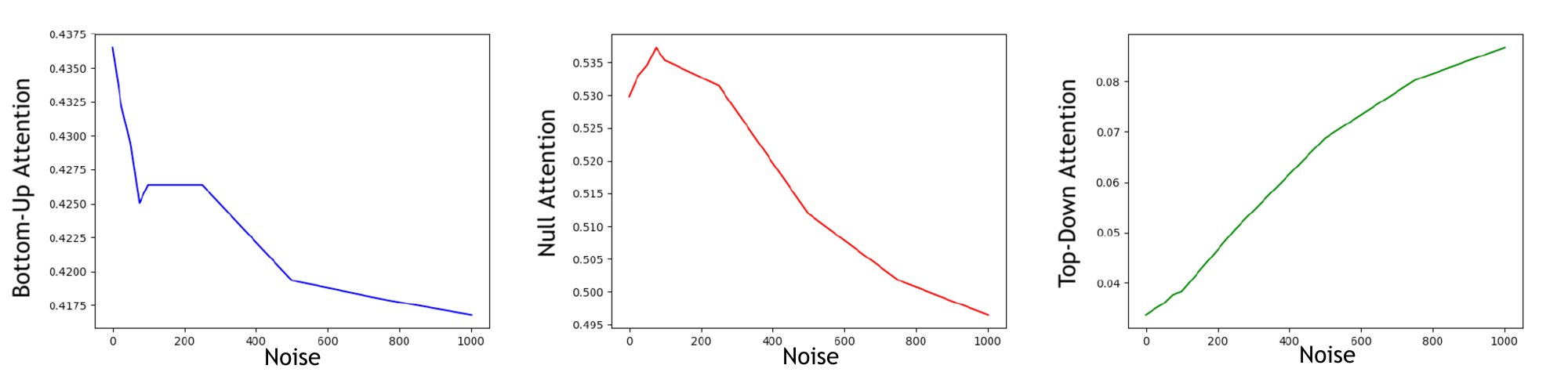}
    %\vspace{1mm}
    \includegraphics[scale=0.25]{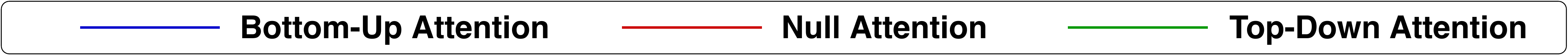}
    \caption{On sequential CIFAR-10, we evaluate on 32x32 test images and change a certain number of the pixels to random values.  We show the \textbf{average activation weight (y-axis)} on Input (left), Null (middle) and Higher Layer (right) plotted against the \textbf{number of random pixels (x-axis)} (total number of pixels = 1024).  We see that as more pixels are set to random values, the model becomes increasingly reliant on the higher-level information.  }
    \label{fig:noise-cifar}
\end{minipage}
\vspace{-2mm}
\end{figure*}
%\subsection{Updating the Recurrent State}

%The overall update consists of first computing the selective activation $\mathcal{S}^l_t$, and then doing the communication between a particular layer, followed by the communication within layers:
%\begin{align}
%    \begin{split}
%            \textbf{h}^l_{t,k} &= G^l_k(\hat{\textbf{h}}^l_{t}) \\
%            \hat{\textbf{h}}^l_{t,k} &= F_k^l(\textbf{h}^l_{t-1,k}, \textbf{h}^{l-1}_{t}) 
%    \end{split} & k \in \mathcal{S}^l_t\label{rim_hier}\\
%    \textbf{h}^l_{t,k} &= \textbf{h}^l_{t-1,k} & k \notin \mathcal{S}^l_t 
%\end{align}
%Note that this choice introduces module level dependence to the recurrent dynamics $G \circ F$. This dependence is constructed using attention mechanisms both between modules in different layers (Communication between different layers $F^l_k$) as well as modules within each layer (Communication within Layer $G_l$).  
%\vs{1}
\subsection{Training}
%\vs{1}

The architecture  proposed here doesn't rely on any additional losses and  thus can be used as a drop-in substitute for LSTMs and GRUs. For training we consider task-specific losses which range from classification and video prediction losses to RL losses depending on the problem.

\section{Related Work}

\textbf{Deep Boltzmann Machines:} Deep Boltzmann machines \citep{salakhutdinov2009deep} resemble our motivating intuitions, as they are undirected and thus have both top-down and bottom-up feedback through an energy function. Additionally, the model captures a probability distribution over the hidden states, such that the role of top-down feedback becomes most important when the bottom-up signal has the most uncertainty. However, a significant problem in deep Boltzmann machines, and with undirected graphical models in general, is that sampling and inference are both difficult. Sampling generally requires an iterative procedure, which may converge slowly. % Additionally blocked Gibbs sampling mixes between modes more slowly as the number of layers grow \citep{bengio2013better, zhao2017hierarchy}.  

\textbf{Helmholtz Machine:} Like autoencoders, the Helmholtz
machine \citep{dayan1995helmholtz} contains two separate networks: a generative network and a
discriminative network, and the wake-sleep
learning algorithm is applied to discover a latent representation
of the data in an unsupervised way. However, unlike in our proposal, the top-down path only
influences the learning of the bottom-up path, and not its actual
current computation.

\textbf{Transformers:} The Transformer architecture \citep{vaswani2017attention} eschews the use of a recurrent state in favor of using attention to pass information between different positions. It lacks an explicit inductive bias towards a bottom-up followed by top-down flow, since at the lowest-layers of the network, the positions in the distant past can only undergo a small amount of processing before they can influence the lower levels of processing of later inputs.  

\textbf{Modulating Visual Processing with Language:} \cite{devries2017visual} proposes to modulate a visual image processing pipeline by adding in language representations at multiple layers, including early layers.  Since the language is already quite abstract, this can be seen as top-down modulation on the model's bottom-up processing.  While this improved results, it assumes that the raw data is associated with text which provides a top-down signal, as opposed to our proposed technique, in which the top-down signal is learned end to end from raw data and dynamically applied using top-down attention. 

\textbf{Hierarchical Multiscale Recurrent Neural Network:} This paper explored an architecture in which the higher layer at previous steps is given as additional inputs to an RNN's lower layers \citep{chung2016hierarchical}. Additionally the higher level information is flushed, copied, or updated, using a sparse gating operation. However a key difference from our work is that we use attention to determine whether and what to read from the higher or lower levels, as opposed to concatenating into the input. Additionally, our approach uses a modular recurrent state.

\textbf{Generative Classifiers:} These classify by selecting a class y to maximize $p(x|y)p(y)$.  This is a top-down form of classification, since a key factor is estimating the likelihood of a given sample $x$ under a generative model $p(x|y)$.  Notably, this purely top-down approach often struggles as learning a model of $p(x|y)$ is generally much harder and much more complex than a direct model of $p(y|x)$ \citep{genclass2002ng}.  

\textbf{Link to Global Workspace Theory:}
The GWT~\citep{baars1997theatre,Dehaene-et-al-2017} proposes that this high-level conscious content is broadcast across the brain via top-down connections
forming an active circuit with the bottom-up connections.
What has been missing from RIMs and which we propose
to study here is therefore a notion of hierarchy
of levels enabling this kind of bidirectional information flow
mediated by attention. One major difference between the current work, and GWT is that,  the kinds of modules that are usually considered in GWT are pretty high level, e.g., face recognition, gait recognition, object recognition, visual routines etc i.e., GWT considers \textit{macro} modules, while the current work focuses more on micro modules.

\section{Experiments}

The main goal of our experiments is to explore the effect of BRIMs on generalization.  In particular, we primarily focus on tasks where the training and testing distribution differ systematically, as we believe that the improved top-down modulation of BRIMs will help the model to adapt to variations unseen during training.  Additionally, we provide ablations and analysis which explore how the different components of the BRIMs architecture affect results.  

We show that BRIMs improve out-of-distribution generalization over sequence length, on sequential MNIST and CIFAR classification, and moving MNIST generation.  We then show that BRIMs better reason about physical movement in a synthetic bouncing balls dataset.  In all of these tasks, the test distribution differs dramatically from the training distribution.  We also show that BRIMs improve results on both language modeling and reinforcement learning. As the data distribution implicitly changes in reinforcement learning as a result of the policy changing during training, this provides further evidence that BRIMs improve out-of-distribution generalization in complex settings.  

%Empirically we will show the following things:-

%\begin{itemize}
%    \item Modular systems can evolve by change in one module at a time, or by duplication and mutation of modules, without risking loss of function in modules that are already well adapted. Well-adapted modules thus represent stable intermediate states such that further evolution of other modules does not jeopardize function of the entire system. This robustness represents a major advantage for any system evolving under changing or competitive selection criteria, and this may explain the widespread prevalence of modular architectures across a very wide range of information processing systems.
%    \item Languags have a natural heirarchical strucutre, by experiments we will show that the proposed model helps in language modelling.
%    \item We also show that the proposed method helps to learn a better predictive model
%    by predicting the motion of balls in bouncing balls, as well as better generalization to out of distribution.
%    \item Finally, we will show that incorporating top-down information helps in learning sample efficient policies for Reinfrocement learning.
%\end{itemize}

\begin{table}
\vs{2}
\begin{center}
\renewcommand{\arraystretch}{1.0}
\resizebox{\columnwidth}{!}{
\begin{tabular}{ ccccc } 
 \hline
 \textit{Algorithm} & \textit{Properties} & \textit{16$\times$16} & \textit{19$\times$19} & \textit{24$\times$24}\\ 
 \hline
 LSTM & --- &86.8 & 42.3 & 25.2 \\
 LSTM & \textsc{h} & 87.2 & 43.5 & 22.9 \\
 LSTM & \textsc{h+b} & 83.2 & 44.4 & 25.3 \\
 LSTM & \textsc{h+a} & 84.3 & 47.5 & 31.0\\
 LSTM & \textsc{h+a+b} & 83.2 & 40.1 & 20.8\\
 \hline
 RMC & \textsc{a} & 89.6 & 54.2 & 27.8\\
 Transformers & \textsc{h+a+b} & \highlight{91.2} & 51.6 & 22.9\\
 RIMs & \textsc{a+m} & 88.9 & 67.1 & 38.1 \\
 \hline
 Hierarchical RIMs & \textsc{h+a+m} & 85.4 & 72.0 & 50.3 \\
 MLD-RIMs & \textsc{h+a+m} & 88.8 & 69.1 & 45.3 \\
 BRIMs (ours) & \textsc{h+a+b+m} & 88.6 & \highlight{74.2} & \highlight{51.4} \\
 \hline
\end{tabular}
}
\end{center}
\caption{Performance on the \textbf{Sequential MNIST resolution generalization:} Test Accuracy \% after 100 epochs.  All models were trained on 14x14 resolution but evaluated at different resolutions; results averaged over 3 different trials.}
\label{table:mnist}
\vspace{-6mm}
\end{table}

\subsection{Baselines}

In our experiments, we consider multiple baselines which use some aspects of BRIMs, such as stacked recurrent layers and modularity. We compare our architecture against various state-of-the-art models (Transformers \citep{vaswani2017attention}, Relational RNNs \citep{santoro2018relational}) while also providing ablation studies based on an LSTM backbone \citep{hochreiter1997long}.  By outperforming these baselines, we demonstrate the necessity of the contributions introduced in the BRIMs architecture. In particular, we try to disentangle the contributions of 
attention [\textsc{a}], hierarchy [\textsc{h}], modularity [\textsc{m}], and bidirectional [\textsc{b}] structure.

\textit{LSTM variants}: We consider variants of LSTMs based on different subsets of important properties. In particular, we experimented with standard LSTMs as well as hierarchical LSTMs without feedback [\textsc{h}], with feedback [\textsc{h+b}],
with attention [\textsc{h+a}], and with both [\textsc{h+a+b}].

\textit{Transformers} [\textsc{h+a}]: Self-attention based multi-layer architecture \citep{vaswani2017attention}. 

\textit{RMC, a relational RNN} [\textsc{a}]: Memory based recurrent model with attention communicating between saved memory and hidden states \citep{santoro2018relational}.  

\textit{Recurrent Independent Mechanisms (RIMs)} [\textsc{a+m}]: Modular memory based single layered recurrent model with attention modulated input and communication between modules \citep{goyal2019recurrent}. 

\textit{Hierarchical RIMs} [\textsc{h+a+m}]: Two layered modular architecture with attention based communication between the modules in the same layer and between different layers. Here the flow of information is unidirectional i.e information only flows from bottom to top, and hence no top-down information is being used. 

\textit{MLD-RIMs, RIMs with Multilayer Dynamics} [\textsc{h+a+m}]: The same as hierarchical RIMs except that the activation of RIMs on the higher layer is copied from the lower layer instead of being computed at the higher layer.  This more tightly couples the behavior of the lower and higher layers.  %. \color{red}(Refer to \hyperref[complex-blocks]{Section ~\ref{complex-blocks}})\color{black} 

\textit{BRIMs} [\textsc{h+a+m+b}]: Our model incorporates all these ingredients: layered structure, where each layer is composed of modules, sparingly interacting with the bottleneck of attention, and allowing top down information flow. 

For more detailed description about the baselines, we refer the reader to section \ref{appendix:baseline_desc} in Appendix.

% \vspace{-3mm}

% \paragraph{LSTM} - One layered LSTM.
% \vspace{-4mm}
% \paragraph{LSTM (H)} - Two layered LSTM.
% \vspace{-4mm}
% \paragraph{LSTM (H + B)} - Two layered LSTMs with top down connections.
% \vspace{-4mm}
% \paragraph{LSTM (H + A)} - Two layered LSTMs with attention modulated recurrent connections between layers.
%  \vspace{-4mm}
%  \paragraph{LSTM (H + A + B)} - Two layered LSTMs with attention modulating both top down as well as bottom up connections.
%  \vspace{-4mm}
% \paragraph{Transformers} - Self-attention based architecture.
% \vspace{-4mm}
% \paragraph{Relational RNN (RMC)} - Memory based recurrent model with attention communicating between saved memory and hidden states.
% \vspace{-4mm}
% \paragraph{Recurrent Independent Mechanisms (RIMs)} - Modular memory based single layered recurrent model with attention modulated input and communication between modules \citep{goyal2019recurrent}. RIMs can be considered as \textbf{(A + M)}.
% \vspace{-4mm}
% \paragraph{Hierarchical RIMs} -  Two layered modular architecture with attention based communication between the modules in the same layer and between different layers. Here the flow of information is unidirectional i.e information only flows from bottom to top, and hence no top-down information is being used. This can be considered as \textbf{(H + A + M)}.
% \vspace{-3mm}
% \paragraph{Complex Hierarchy} - Similar to hierarchical RIMs but with each module spanning two layers. Can be considered as \textbf{(H + A + M)}. (Refer \hyperref[complex-blocks]{Section 6.4} for details)

\begin{table}
\vs{3}
\begin{center}
\renewcommand{\arraystretch}{1.0}
\resizebox{\columnwidth}{!}{
\scalebox{0.95}{\begin{tabular}{ ccccc } 
 \hline
 %\textbf{Algorithm} & \textbf{19 x 19} & \textbf{24 x 24} & \textbf{32 x 32} \\ 
 \textit{Algorithm} & \textit{Properties} & \textit{19$\times$19} & \textit{24$\times$24} & \textit{32$\times$32} \\ 
 \hline
 LSTM & ---  & 54.4 & 44.0 & 32.2\\
 LSTM & \textsc{h} & 57.0 & 46.8 & 33.2\\
 LSTM & \textsc{h+b} & 56.5 & 52.2 & 42.1 \\
LSTM & \textsc{h+a} & 56.7 & 51.5 & 40.0 \\
LSTM & \textsc{h+a+b} & 59.9 & 54.6 & 43.0\\
\hline
 RMC & \textsc{a} & 49.9 & 44.3 & 31.3 \\
 RIMs & \textsc{a+m} & 56.9 & 51.4 & 40.1 \\
 \hline
% RMC & & & \\
Hierarchical RIMs & \textsc{h+a+m} & 57.2 & 54.6 & 46.8 \\
MLD-RIMs & \textsc{h+a+m} & 56.8 & 53.1 & 44.5\\
%BRIMs (ours) & \textbf{60.1} & \textbf{57.7} & \textbf{52.2} \\
BRIMs (ours) & \textsc{h+a+b+m} & \highlight{60.1} & \highlight{57.7} & \highlight{52.2} \\
 \hline
\end{tabular}}
\vspace{-4mm}
}
\end{center}
\caption{Performance on \textbf{Sequential CIFAR generalization:} Test Accuracy \% after 100 epochs.  Both the proposed and the Baseline model (LSTM) were trained on 16x16 resolution but evaluated at different resolutions; results averaged over 3 different trials.}
\label{table:cifar}
\vspace{-4mm}
\end{table}

\subsection{Model Setup}
We use a 2-layered setup with each layer consisting of a set of modules. The proposed architecture has 2 degrees of freedom:  the number of modules in each layer and number of active modules in each layer. For the supervised loss, the final state of the recurrent model is passed through a feed-forward network before making a prediction. For RL experiments, the concatenation of the state of lower level modules as well as higher level modules is used to compute the policy. For video prediction experiments, the state of the lower level module is fed into the decoder, and used for generating the image. Unless otherwise indicated we always  (a) learn an embedding for input tokens before feeding it to the RNNs, (b) use adam with a learning rate
of 0.001 and momentum of 0.9. We include more experimental results in the supplementary material.

%\subsection{Adding Task}
%This is an example of a task where complex reasoning is involved. In order to be able to tackle this task well, the recurrent models need to learn the concept of addition. We model the task of addition by considering it as a $T$ length sequential problem with the input as $\{ (x_t, a_t, b_t) \}_{t=1}^T$. Here $x_t$ denotes a number drawn uniformly randomly between $0$ and $1$. $a_t$ is a one-hot vector that denotes the index of the first element to consider in addition while $b_t$ denotes the second element. Thus, if $a_t = 1$ at $t=t_i$ and $b_t = 1$ at $t=t_j$, then the task boils down to estimating $x_{t_i} + x_{t_j}$. For our setting, we train the models on sequence length of 500 and then test on sequence length 1000 in an attempt to understand which models are able to generalize well.

\begin{figure}
  \centering
  {\includegraphics[width=0.75\linewidth]{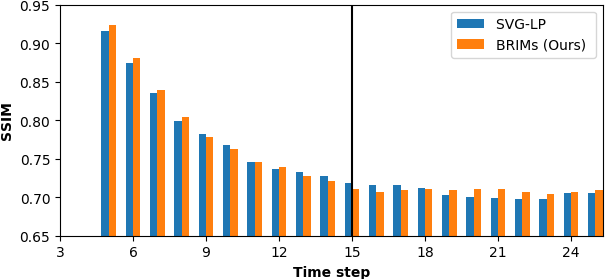}}
  %\\ \subfloat (a) \\
  {\includegraphics[width=0.75\linewidth]{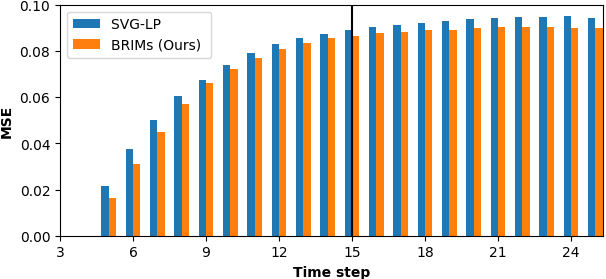}}
  %\\ \subfloat (b)
  \caption{Comparison of BRIMs and SVG-LP
  \citep{denton2018stochastic} 
  on Stochastic Moving MNIST, evaluated using (a) SSIM (higher is better) and (b) MSE (lower is better).
  Models are trained on digits 0-7 and tested with digits 8-9.
  The models are conditioned on 5 time steps, and trained for 10 more time steps. SSIM is calculated by rolling out the models into future time steps.}
  \label{fig:mnist}
  \vspace{-5mm}
\end{figure}
%\vs{5}
\subsection{Sequential MNIST and CIFAR}
\vs{1}
\label{sPixel}

These tasks involve feeding a recurrent model the pixels of an image in a scan-line order and
producing a classification label at the end of the sequence. To study the generalization capacity of different recurrent models, we train on sMNIST at a resolution of (14,14) and test on the resolutions (16,16), (19,19) and (24,24). Similarly, we perform training for sCIFAR at resolution (16,16) and testing on (19,19), (24,24) and (32,32).  Because these images contain many distracting and uninformative regions, a model which is better able to ignore these patterns should generalize better to longer sequences.  Additionally many details in these images are difficult to understand without having good top-down priors.  These will be most salient when the specific character of these details changes as, for example, when we increase the resolution.  See Tables~\ref{table:cifar},~\ref{table:mnist} for results and appendix section~\ref{appendix:mnist} for more details.

\paragraph{Analysis of Top Down Attention.} We perform analysis on learned BRIMs models to demonstrate the nature of the learned bidirectional flow on sCIFAR task. We show that the model's reliance on top-down information increases when the bottom-up information is less reliable. Further, in general, the models query higher level top-down information only sparingly.

We experimentally verified that almost every example attends to the higher level, yet only a small fraction of steps attend to the higher level on typical examples.  Quantitatively, 95.1\% of images accessed the higher level at least five times (with a score $\geq$ 50\%), while on average only 2.86\% of the total attention (averaged over all modules) was to the higher level. This shows that the model uses top-down information consistently but sparsely.

We also analysed the influence of top-down connections when there is noise in the data. We show that if we train normally but make some of the pixels random (uniformly random noise) at test time, then the model puts more of its attention on the higher level, as opposed to the input (Figure~\ref{fig:noise-cifar}). This is evidence that the model learns to rely more heavily on expectations and prior knowledge when the input sequence is less reliable.

\begin{figure}
    \centering
    \includegraphics[width=0.9\linewidth,trim={0.5cm 0 0.5cm 2cm},clip]{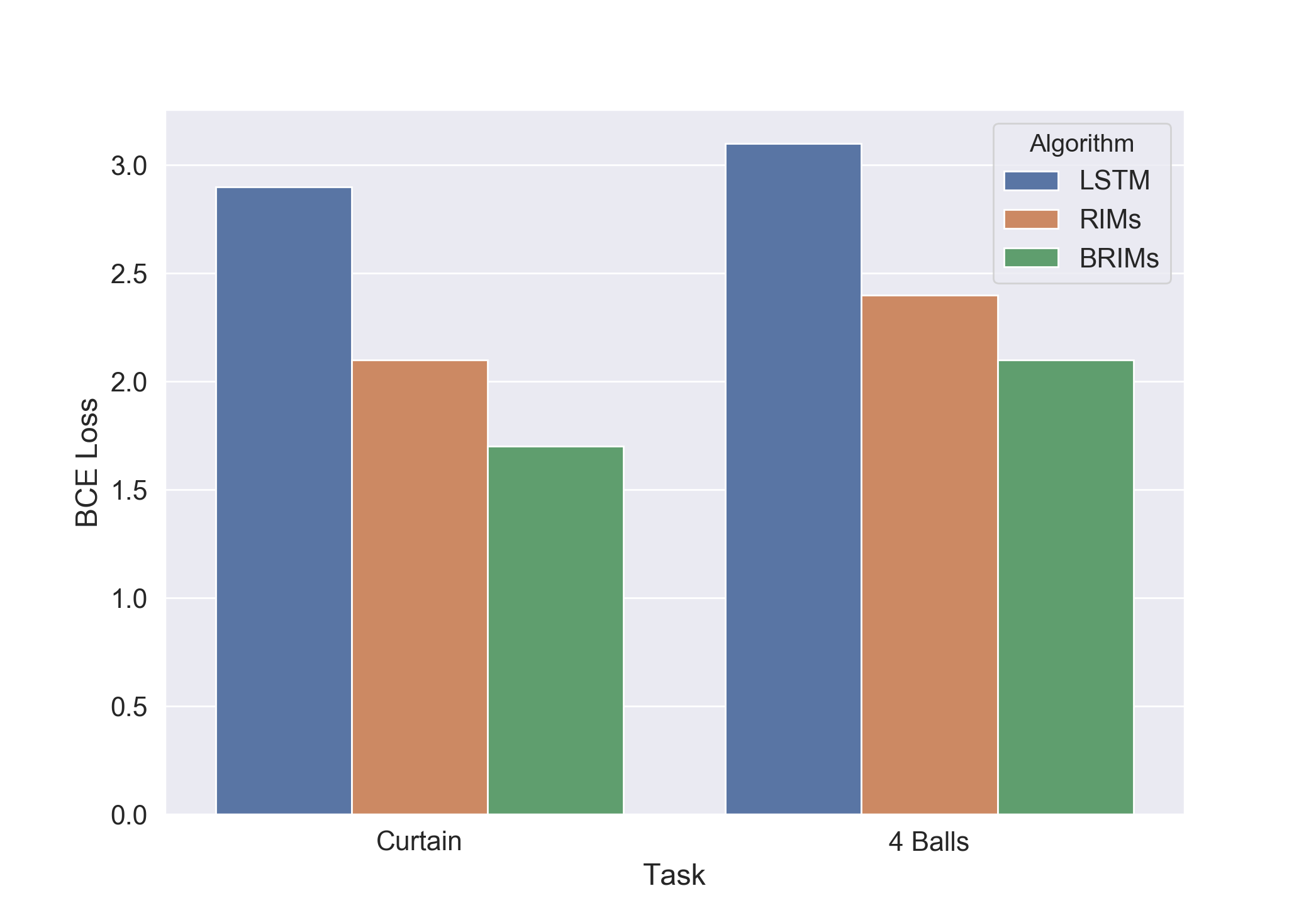}
    \vspace{-4mm}
    \caption{We study the performance of the proposed model as compared to the LSTM as well as RIMs baseline. We feed the ground truth for the first 15 time steps, and then we ask the model to predict the next 35 time steps. During rollout phase, the proposed model performs better in terms of accurately predicting the dynamics of the balls for the 4Balls scenario, as well as for the more difficult occlusion scenario, as reflected by the lower BCE.}
    \label{fig:bouncing_balls_loss}
    \vspace{-4mm}
\end{figure}

\subsection{Adding Task}
\vs{1}

In the adding task we consider a stream of numbers as inputs (given as real-values) and then indicate which two numbers should be added together as two additional input streams which vary randomly between examples.  The length of the input sequence during testing (2000) is longer than during training (1000).  This is a simple test of the model's ability to ignore the distractor numbers. While we train on two number additions, we also test it on cases where more than two numbers are to be added. We provide the results in Table \ref{table:adding}. We also point the readers to appendix section~\ref{appendix:adding} for additional experiments and details regarding this setup. Here, our Random Prediction baseline, unlike the models, assumes knowledge of the number of digits to add.

\vs{1}
\subsection{Moving MNIST: Video Prediction}
\vs{1}
\label{movingMNIST}

We use the Stochastic Moving MNIST (SM-MNIST)
dataset introduced by \citet{denton2018stochastic} which consists of sequences of frames of size 64 x 64,
containing one or two MNIST digits moving and bouncing
off the edge of the frame (walls). Training sequences were generated on the fly by sampling
two different MNIST digits from the training set (60k total
digits). We trained the proposed model on SM-MNIST by conditioning on 5 frames and training
the model to predict the next 10 frames in the sequence, and compared it to the SVG-LP model \citep{denton2018stochastic}.

To test out-of-distribution generalization capability, we trained models on image sequences from the SM-MNIST dataset containing only digits 0-7, and tested them on sequences containing only digits 8 \& 9. We compared the Structural Similarity index (SSIM) and the Mean Squared Error (MSE) of the best generated sequences from the baseline SVG-LP model, and those from ours. We followed the same evaluation protocol as SVG, and the results in \hyperref[fig:mnist]{Figure~\ref{fig:mnist}} demonstrate that our proposed method performs consistently better. For details about the experimental setup, we refer the reader to section \ref{appendix:moving_appendix} in Appendix.

\paragraph{Analysis of Top Down Attention.} We hypothesize that higher level encodes top-down beliefs (eg. location of digits) that drive the lower level to draw the digits for the next frame. To support this claim, we train a classifier to predict both the digits’ locations from the learned representation in 1 layered RIMs (82\% accuracy), higher layer of 2 layered LSTM (76\% accuracy), higher layer of BRIMs (88\% accuracy) and lower layer of BRIMs (58\% accuracy). This indicates that the information about the positions of the digits is stored in higher levels that could be useful for better modelling of the dynamics at a lower level especially when the digits overlap.

\begin{table}[t!]
\centering
\renewcommand{\arraystretch}{1.0}
\resizebox{\columnwidth}{!}{
\begin{tabular}{c|c|c|c} 
 \toprule
    \textit{Number of Values} & \textit{Random Prediction} & \textit{LSTM} & \textit{BRIMs (ours)} \\
    \midrule
        2 & 0.500 & 0.0842 & \highlight{0.0000} \\
        3 & 1.000 & 0.4601 & \highlight{0.0999} \\
        4 & 1.333 & 1.352 & \highlight{0.4564} \\
        5 & 2.500 & 2.738 & \highlight{1.232} \\
        10 & \highlight{9.161} & 17.246 & 11.90 \\
\bottomrule
\end{tabular}
}
\vs{3}
%\vs{4}
\caption{Results on Adding task with training and testing on 1000 and 2000 sequence lengths respectively. Number of values to be added are varied from 2 to 10 during test time.}
% \caption{We train on sequences of length 1000 where 2 values must be summed.  We test on sequence length 2000 and vary the number of values to be summed between 2 and 10.  We note much better generalization when using BRIMs.}
\label{table:adding}
%\vs{5}
\vspace{-6mm}
\end{table}
% \begin{figure*}[!htp]
%     \centering
%     \includegraphics[width=.9\linewidth]{images/Multiple_SSIM_89_final.png}
%     \caption{SSIM for the SVG-LP model~\citep{denton2018stochastic} and the RIM model trained on Stochastic Moving MNIST images with digits 0-7, and tested on those with digits 8-9. The models were conditioned on 5 time steps, and trained for 10 more time steps. SSIM was calculated by rolling out the models into future time steps.}
%     \label{fig:svg_ssim}
% \end{figure*}

% \begin{figure}
%   \centering
%   {\includegraphics[width=0.75\linewidth]{images/Multiple_SSIM_89_final.png}}
%   %\\ \subfloat (a) \\
%   {\includegraphics[width=0.75\linewidth]{images/Multiple_MSE_89_final.png}}
%   %\\ \subfloat (b)
%   \caption{Comparison of BRIMs and SVG-LP
%   \citep{denton2018stochastic} 
%   on Stochastic Moving MNIST, evaluated using (a) SSIM (higher is better) and (b) MSE (lower is better).
%   Models are trained on digits 0-7 and tested with digits 8-9.
%   The models are conditioned on 5 time steps, and trained for 10 more time steps. SSIM is calculated by rolling out the models into future time steps.}
%   \label{fig:mnist}
%   \vspace{-6mm}
% \end{figure}

\subsection{Handling Occlusion in Bouncing Balls}
%\vs{1}
\label{bouncingballs}

We study the ability of the proposed model to model physical reasoning capabilities, on the bouncing balls task, a standard environment for evaluating physical reasoning capabilities exhibiting complex non-linear physical dynamics. We train BRIMs on sequences of $64 \times 64$ binary images over 51 time-steps that contain four bouncing balls with different masses corresponding to their radii. The balls are initialized with random initial positions, masses and velocities. Balls bounce elastically against each other and the image window. We also test the proposed method  in the scenario when an invisible ``curtain'' \citep{van2018relational} is present, which further tests the degree to which the model correctly understands the underlying dynamics of the scene.  For vizualizations about the predictions from the proposed model, refer to section \ref{appendix:bouncing_balls} in Appendix.

\paragraph{Analysis of Top Down Attention.} Achieving good performance on this task requires the model to track the ball from behind the curtain. We find that whenever the perceptual input is less reliable (i.e a ball is occluded) the perception is relied more by the information coming from higher level layers (higher top-down attention score).

%\vspace{-25mm}

%\vs{1}
\subsection{Language Modelling}
%\vs{1}

We consider word-level language modeling on the WikiText-103 dataset which has been used as a benchmark in previous work \citep{santoro2018relational}.  Language has an organic hierarchical and recurrent structure which is both noisy and flexible, making it an interesting and apt task to study for the proposed model.  Table~\ref{table:wiki} shows Validation and Test set perplexities for various models. We find that the proposed method improves over the standard LSTM as well as RMCs. 
\begin{table}
%\vs{1}
\begin{center}
\renewcommand{\arraystretch}{1.0}
%\resizebox{\columnwidth}{!}{
\begin{tabular}{cccc} 
 \hline
 \textit{Algorithm} & \textit{Properties}  & \textit{Valid} & \textit{Test} \\ 
 \hline
 LSTM & --- & 38.2 & 41.8 \\
 RMC* (ours) & \textsc{h} & 36.2 & 38.3 \\
Hierarchical RIMs & \textsc{h+a+m} & 36.1 & 38.1 \\
BRIMs (ours) & \textsc{h+a+b+m} & \highlight{35.5} & \highlight{36.8} \\
 \hline
\end{tabular}
%}
\end{center}
%\vs{3}
\caption{Perplexities on the WikiText-103 dataset. 
$(*)$ refers to our implementation of the model. The actual results from RMC are even better as compared to the proposed model, but the idea here is not to beat well tuned state of the art methods.}
\label{table:wiki}
\vspace{-5mm}
\end{table}

% \begin{figure}
%     \centering
%     \includegraphics[width=0.9\linewidth,trim={0.5cm 0 0.5cm 2cm},clip]{figures/bouncing_balls_loss.png}
%     \vspace{-4mm}
%     \caption{We study the performance of the proposed model as compared to the LSTM as well as RIMs baseline. We feed the ground truth for the first 15 time steps, and then we ask the model to predict the next 35 time steps. During rollout phase, the proposed model performs better in terms of accurately predicting the dynamics of the balls for the 4Balls scenario, as well as for the more difficult occlusion scenario, as reflected by the lower BCE.}
%     \label{fig:bouncing_balls_loss}
%     \vspace{-4mm}
% \end{figure}
\vspace{-4mm}
\subsection{Reinforcement Learning}
%\vs{1}

Top-down and bottom-up signals are clearly differentiated in the case of reinforcement learning (or in active agents, more generally) in that they lead to different optimal behavior.  Since top-down signals change more slowly, they often suggest long-term and deliberative planning.  On the other hand, bottom-up signals which are surprising can often require an immediate reaction.  Thus, in selecting actions, points in which the top-down system dominates may benefit from activating a system with a long-term plan, whereas points where the bottom-up information dominates could benefit from a system which immediately takes action to resolve an unexpected threat.  On the whole, we expect that dynamically modulating activation based on bottom-up and top-down signal importance will be especially important in reinforcement learning.  Additionally, in reinforcement learning the data distribution implicitly changes as the policy evolves over the course of training, testing the model's ability to correctly handle distribution shift.  

To investigate this we use an RL agent trained using Proximal Policy Optimization (PPO) \citep{schulman2017proximal} with a recurrent network producing the policy. We employ an LSTM as well as RIMs architecture \citep{goyal2019recurrent} as baselines and compare results to the proposed architecture. This was a simple drop-in replacement and did not require changing any of the hyperparameters for PPO. We experimented on 10 Atari games (chosen alphabetically) and ran three independent trials for each game (reporting mean and standard deviation).  We found that simply using BRIMs for the recurrent policy improves performance (\hyperref[table:atari]{Table~\ref{table:atari}}).  We also found an improvement on Ms. Pacman, which requires careful  planning and is not purely reactive. 

\begin{table}[t!]
\large
\renewcommand{\arraystretch}{1.2}
%\vs{1}
\centering
\resizebox{\columnwidth}{!}{
\begin{tabular}{c|rm{1em}l|rm{1em}l|rm{1em}l}
    \toprule
    \textit{Environment} & \multicolumn{3}{c}{\textit{LSTM}} & \multicolumn{3}{c}{\textit{RIMs}} &  \multicolumn{3}{c}{\textit{BRIMs (ours)}}\\
    \midrule
    Alien & 1612  &$\pm$& 44 & 2152 &$\pm$& 81 & \highlight{4102} &$\pm$& 400 \\
    Amidar & 1000  &$\pm$& 58 & 1800 &$\pm$& 43 & \highlight{2454} &$\pm$& 100\\
    Assault &  4000 &$\pm$& 213 & 5400 &$\pm$& 312 & \highlight{5700} &$\pm$& 320 \\
    Asterix & 3090 &$\pm$& 420 & 21040 &$\pm$& 548 & \highlight{30700} &$\pm$& 3200\\
    Asteroids & 1611 &$\pm$& 200 & \highlight{3801} &$\pm$& 89 & 2000 &$\pm$& 300\\
    Atlantis & 3.28M &$\pm$& 0.20M & 3.5M &$\pm$& 0.12M & \highlight{3.9M} &$\pm$& 0.05M \\
    BankHeist & 1153 &$\pm$& 23 & \highlight{1195} &$\pm$& 4 & 1155 &$\pm$& 20 \\
    BattleZone & 21000 &$\pm$& 232 & 22000 &$\pm$& 324 & \highlight{25000} &$\pm$& 414 \\
    BeamRider & 698 &$\pm$& 100 & \highlight{5320} &$\pm$& 300 & 4000 &$\pm$& 323\\
    MsPacMan & 4598 &$\pm$& 100 & 3920 &$\pm$& 500 & \highlight{5900} &$\pm$& 1000 \\
\bottomrule
%\vspace{-3mm}
\end{tabular}
}
\caption{Results on the Atari Reinforcement learning task using PPO with different recurrent policy networks.}
\label{table:atari}
\vspace{-4mm}
\end{table}

\section{Ablations and Analysis}
We have demonstrated substantially improved results using BRIMs.  However, it is essential to understand which aspects of the newly proposed model are important. To this end we conduct ablation studies to demonstrate the necessity of bidirectional structure, attention, modularity, and hierarchy.  

\vspace{-1mm}
\paragraph{Role of Bidirectional Information Flow.}
\label{td}
We demonstrate that in different architectures, having top down information flow helps in generalization. Tables \ref{table:cifar},~\ref{table:mnist},~\ref{table:wiki} demonstrate the effectiveness of using top down connections in standard LSTMs, attention based LSTMs and hierarchical modules. We maintain that higher layers are able to filter out important information from lower layers and thus conditioning on it provides much more context as to what is relevant at any time in a sequence.  However, the results still fall short of BRIMs, indicating that this component alone is not sufficient. 

% \subsection{Consistent Utilization of the Higher Level}
% We also investigated the nature of the learned bidirectional flow on sequential CIFAR-10 classification. The lower level can  dynamically choose whether to use information from input or the higher level.
% The lower level should query the higher level only if the latter contains information which might be relevant for solving the task at hand. We can analyze the frequency with which the lower level queries information from the higher level, i.e., whether it exploits top-down information.

% We experimentally verified that almost every example attends to the higher level, yet only a small fraction of steps attend to the higher level on typical examples.  Quantitatively, 95.1\% of images accessed the higher level at least five times, while on average only 2.86\% of the total attention (averaged over all modules) was to the higher level.  

\vspace{-1mm}
\paragraph{Role of Attention.}
Tables~\ref{table:cifar},~\ref{table:mnist},~\ref{table:wiki},~\ref{table:atari} and Figure~\ref{fig:mnist} show the importance of having attention, as Transformers, RMC, attention-modulated LSTMs, RIMs and BRIMs lead to much better performance and generalization in domains ranging from supervised learning to reinforcement learning than their non attention counterparts. This suggests that key-value attention mechanism is better able to extract out important information and dependency as opposed to standard connections.  

\vspace{-1mm}
\paragraph{Role of Modularity.}
Tables~\ref{table:cifar},~\ref{table:mnist},~\ref{table:wiki},~\ref{table:atari} indicate that having layers as a composition of smaller modules with their own independent dynamics outperform architectures without modular decomposition. In particular, we note that RIMs, their hierarchical variants and BRIMs perform much better than their LSTM counterparts. This shows that having independent dynamics for modules as well as a communication channel between them are important ingredients for generalization.
\vspace{-1mm}
\paragraph{Role of Hierarchy.}
We consistently found that models with hierarchical structure perform much better than their shallow counterparts (Tables ~\ref{table:cifar},~\ref{table:mnist},~\ref{table:atari}). This relates to the discussion in Section \hyperref[td]{7.1} that higher layers are able to abstract and filter out important information and concepts learned in lower layers.

\vspace{-2mm}
\paragraph{Capacity of Network.}
While BRIMs introduce additional parameters through the attention mechanisms, they are roughly cancelled out by the block diagonal structure in the independent module dynamics. As an example, our LSTM and BRIMs models for sCIFAR both contain $\sim$ 1.5M parameters. Our additional experiments on larger LSTMs showed a drop in performance (2-3\%) on sMNIST and sCIFAR.

%\vspace{-2mm}

%The consciousness prior~\citep{bengio2017consciousness} attempts
%to provide a machine learning motivation for the Global Workspace Theory or GWT~\citep{baars1997theatre,Dehaene-et-al-2017}. This prior states
%that high-level variables have a joint distribution which can be
%captured by a sparse factor graph, with the same parameterized computations (factors, in the factor graph nomenclature) being applicable (like rules) to many possible instances of %high-level random variables. Inference
%in this sparse factor graph would naturally be performed by consideringonly a few elements at the time, which would correspond to the current conscious content. 
% \section{Discussion}

% \paragraph{Link to Global Workspace Theory.}
% The GWT~\citep{baars1997theatre,Dehaene-et-al-2017} proposes that this high-level conscious content is broadcast across the brain via top-down connections
% forming an active circuit with the bottom-up connections.
% What has been missing from RIMs and which we propose
% to study here is therefore a notion of hierarchy
% of levels enabling this kind of bidirectional information flow
% mediated by attention. One major difference between the current work, and GWT is that,  the kinds of modules that are usually considered in GWT are pretty high level, e.g., face recognition, gait recognition, object recognition, visual routines etc i.e., GWT considers \textit{macro} modules, while the current work focuses more on micro modules.

\section{Conclusion}

Top-down and bottom-up information are both critical to robust and accurate perception.  
%How to combine these signals has been a central topic of focus within not just deep learning, but also the entire field of cognitive science for several decades.  
Exploring the combination of these signals in network architectures has an important history in deep learning, and has been a central question in cognitive science for several decades.
Our work focuses on the idea that attention can be used to give models explicit dynamic and context dependent control over the combination of top-down and bottom-up signals.  Our simulation experiments have shown a critical role for network modularity: ensuring that specific top-down signals combine with specific bottom-up signals.  Using these insights we have proposed a new algorithm, \emph{Bidirectional Recurrent Independent Mechanisms (BRIMs)}, which achieves substantial improvements on sequence-length generalization tasks, language modeling, video generation, and reinforcement learning on Atari.  A detailed ablation study gives evidence that combining top-down and bottom-up signals through attention is a critical component for the improved results.  

\section*{Acknowledgements}
The authors acknowledge the important role played by their colleagues at Mila throughout the duration of this work. The authors  would like to thank Sergey Levine and Nasim Rahaman for useful discussions. The authors would also like to thank Mark Pickett for feedback on the earlier version of the paper. The authors are grateful to NSERC, CIFAR, Google, Samsung, Nuance, IBM, Canada Research Chairs, Canada Graduate Scholarship Program, Nvidia for funding, and Compute Canada for computing resources.

\bibliography{main}
\bibliographystyle{icml2020}

\newpage
~\\
\newpage

\appendix
\onecolumn

\section{Appendix}
\subsection{Baseline Description}
\label{appendix:baseline_desc}
In our experiments, we consider multiple baselines which use some aspects of BRIMs, such as stacked recurrent layers and modularity. We compare our architecture against various state of the art models (Transformers \citep{vaswani2017attention}, Relational RNNs \citep{santoro2018relational}) while also providing ablation studies based on LSTM \citep{hochreiter1997long} backbone.  By outperforming these baselines, we demonstrate the necessity of the contributions introduced in the BRIMs architecture. In particular, we try to disentangle the contributions of 
attention [\textsc{a}], hierarchy [\textsc{h}], modularity [\textsc{m}], and bidirectional [\textsc{b}] structure.

\textit{LSTM variants}: We consider variants of LSTMs based on different subsets of important properties. In particular, we experimented with standard LSTMs as well as hierarchical LSTMs without feedback [\textsc{h}], with feedback [\textsc{h+b}],
with attention [\textsc{h+a}], and with both [\textsc{h+a+b}].

\textit{Transformers} [\textsc{h+a}]: Self-attention based multi-layer architecture \citep{vaswani2017attention}. 

\textit{RMC, a relational RNN} [\textsc{a}]: Memory based recurrent model with attention communicating between saved memory and hidden states \citep{santoro2018relational}.  

\textit{Recurrent Independent Mechanisms (RIMs)} [\textsc{a+m}]: Modular memory based single layered recurrent model with attention modulated input and communication between modules \citep{goyal2019recurrent}. 

\textit{Hierarchical RIMs} [\textsc{h+a+m}]: Two layered modular architecture with attention based communication between the modules in the same layer and between different layers. Here the flow of information is unidirectional i.e information only flows from bottom to top, and hence no top-down information is being used. 

\textit{MLD-RIMs, RIMs with Multilayer Dynamics} [\textsc{h+a+m}]: The same as hierarchical RIMs except that the activation of RIMs on the higher layer is copied from the lower layer instead of being computed at the higher layer.  This more tightly couples the behavior of the lower and higher layers.  %. \color{red}(Refer to \hyperref[complex-blocks]{Section ~\ref{complex-blocks}})\color{black} 

\textit{BRIMs} [\textsc{h+a+m+b}]: Our proposed model incorporates all these ingredients: layered structure, where each layer is composed of modules, sparingly interacting with the bottleneck of attention, and allowing top down information flow. 

\subsection{sMNIST}
For the baseline LSTM variants, we perform hyperparameter tuning by considering the dimension of hidden state in each layer to be chosen from \{300,600\}. We also experimented with learning rates of 0.001, 0.0007 and 0.0003. For RIMs, Hierarchical RIMs and MLD-RIMs, we experiment with the same hidden state size ( = sum of sizes of all modules of the layer) and learning rate. Number of modules are chosen from the set \{3,5,6\} and the number active at any time is roughly around half of the total number of modules.

We use an encoder to embed the input pixels to a 300 dimensional vector. We consider the lower hidden state to comprise of 6 modules and the higher state of 3 modules. The modules at lower levels are of 50 dimensions while at higher levels they are of 100 dimensions. We maintain 4 modules active at lower level and 2 at higher level at any given time.

We train the model using Adam optimizer with a learning rate of 0.0007. We clip the gradients at 1.0 for stability and train the model with dropout of 0.5 for 100 epochs. Evaluation is obtained according to best performance on a validation split.
\label{appendix:mnist}

\begin{table*}[]
\begin{minipage}{1.0\textwidth}
    \centering
\begin{tabular}{c|cccc}
    Task & Hidden Dimensions & Number of RIMs ($n_l$) & Number of Active RIMs ($m_l$) \\
    \hline
    Sequential MNIST & ( 300 , 300 ) & ( 6 , 3 ) & ( 4 , 2 )\\
    Sequential CIFAR & ( 300 , 300 ) & ( 6 , 6 ) & ( 4 , 4 ) \\
    Moving MNIST &( 300 , 300 )  & ( 6 , 6 )&( 4 , 4 )  \\
    Bouncing Balls &( 300 , 300 ) & ( 5 , 5 ) & ( 3 , 3 ) &\\
    Language Modelling &( 300 , 300 ) &( 6 , 6 ) &( 4 , 4 ) \\
    Reinforcement Learning &( 300 , 300 ) &( 6 , 6 ) &( 4 , 4 ) \\
    Adding & ( 300 , 300 ) & ( 5 , 5 ) & ( 3 , 3 )
\end{tabular}
\caption{Hyperparameters for experiments.  We list the number of hidden units (combined over all RIMs) as well as the number of RIMs on each layer.  }
    \label{tab:my_label}
\end{minipage}
\end{table*}

\subsection{sCIFAR}
\label{appendix:cifar}
For the baseline LSTM variants, we perform hyperparameter tuning by considering the dimension of hidden state in each layer to be chosen from \{300,600\}. We also experimented with learning rates of 0.001, 0.0007 and 0.0003. For RIMs, Hierarchical RIMs and MLD-RIMs, we experiment with the same hidden state size and the same learning rates. We choose the number of modules from the set \{3,5,6\} and the number active at any time is roughly around half of the total number of modules.

We train a model which takes CIFAR images as a sequence of pixels and predicts the class label.  We downsample to 16x16 (a sequence length of 256) during training and use nearest-neighbor downsampling.  

We provide the model with all of the three colors as inputs on each step we use a separate encoder for each channel of the RGB image. We maintain the size of each module at both the lower and higher layer to be 50 and the number of modules in both the layers 6. We restrict the number of activated modules at any time step to be 4 for both the layers. The inputs received by the model are encoded into a 300 dimensional vector.

We train the model using Adam optimizer with learning rate 0.0007. We further clip the gradients at 1.0 in order to stabilize training. We train the model with embedding dropout of 0.5 for 100 epochs and use a validation split to obtain the results on the test set corresponding to best accuracy on validation split.

\subsection{Adding}
For all the variants, we perform hyperparameter search for learning rate from the set \{0.001, 0.0007, 0.0003\}. Apart from learning rate, we didn't perform any other kind of hyperparameter search.

In the adding task we consider a stream of numbers as inputs (given as real-values) and then indicate which two numbers should be added together as a set of two input streams which varies randomly between examples.  The length of the input sequence during testing is longer than during training.  This is a simple test of the model's ability to ignore the numbers which it is not tasked with adding together.  We show that BRIMs provides substantially faster convergence on the adding task. We also evaluate the model's performance on adding multiple numbers even when it is trained on adding only two. We demonstrate that BRIMs generalize better for longer testing sequences as well as when the number of numbers to be added changes between training and evaluation (Tables ~\ref{table:adding_digits} and ~\ref{table:adding}).

For the task we consider a linear encoder that encodes the three input streams into a 300 dimensional vector and a linear decoder that gives the final output given the final hidden state at the higher layer. We use 5 modules, each comprising of a 60 dimensional vector, at both the lower and higher layers. We also maintain that 3 modules remain active at any time at both the lower and higher layers.

The model is trained end to end with a learning rate of 0.001 using Adam Optimizer. We clip the gradients at 0.1 and use dropout of 0.1. We perform two experiments, outlined below:

\begin{itemize}
    \item We train the model to add two numbers from 1000 length sequences and evaluate it on adding variable number of numbers (2 - 10) on 2000 length sequences. (Table~\ref{table:adding})
    \item We train the model to add a mixture of two and four numbers from 50 length sequences and evaluate it on adding variable number of numbers (2 - 10) on 200 length sequences. (Table~\ref{table:adding_digits})
\end{itemize}

\begin{table}
\small
\centering
\resizebox{0.6\columnwidth}{!}{
\begin{tabular}{ c|c c  c } 
 \toprule
    \textit{Number of Values} & \textit{Random Prediction} & \textit{LSTM}  & \textit{BRIMs (ours)} \\
    \midrule
        2 & 0.500 & 0.2032  & 0.0003 \\
        3 & 1.000 & 0.2678  & 0.0002 \\
        4 & 1.333 & 0.3536  & 0.0002 \\
        5 & 2.500 & 0.5505  & 0.0058 \\
        10 & 9.161 & 3.5709  & 2.078 \\
\bottomrule
\end{tabular}
}
%\vs{4}
\caption{We train on sequences with either 2 or 4 values to be summed, and on a sequence length of 50.  We test on a sequence length 200 with different numbers of values to be summed.  This demonstrates that BRIMs improves generalization when jointly varying both the sequence length and the numbers of values to be summed.}
\label{table:adding_digits}
%\vspace{-4mm}
%\vs{5}
\end{table}

% \begin{table}
% \small
% \centering
% \resizebox{0.6\columnwidth}{!}{
% \begin{tabular}{ c|c c c } 
%  \toprule
%     \textit{Number of Values} & \textit{Random Prediction} & \textit{LSTM} & \textit{BRIMs (ours)} \\
%     \midrule
%         2 & 0.500 & 0.0842 & 0.0000 \\
%         3 & 1.000 & 0.4601 & 0.0999 \\
%         4 & 1.333 & 1.352 & 0.4564 \\
%         5 & 2.500 & 2.738 & 1.232 \\
%         10 & 9.161 & 17.246 & 11.90 \\
% \bottomrule
% \end{tabular}
% }
% %\vs{4}
% \caption{We train on sequences of length 1000 where 2 values must be summed.  We test on sequence length 2000 and vary the number of values to be summed between 2 and 10.  We note much better generalization when using BRIMs.  }
% \label{table:adding_length}
% %\vspace{-4mm}
% %\vs{5}
% \end{table}

% \begin{table}
% \small
% \centering
% \resizebox{0.6\columnwidth}{!}{
% \begin{tabular}{ c|c c c } 
%  \toprule
%     \textit{Number of Values} & \textit{Random Prediction} & \textit{LSTM} & \textit{BRIMs (ours)} \\
%     \midrule
%         2 & 0.500 & 0.0077 & 0.0000 \\
%         3 & 1.000 & 0.4172 & 0.0999 \\
%         4 & 1.333 & 1.653 & 0.4564 \\
%         5 & 2.500 & 3.772 & 1.232 \\
%         10 & 9.161 & 22.55 & 11.90 \\
% \bottomrule
% \end{tabular}
% }
% %\vs{4}
% \caption{We train on sequences of length 1000 where 2 values must be summed.  We test on sequence length 2000 and vary the number of values to be summed between 2 and 10.  We note much better generalization when using BRIMs.  }
% \label{table:adding_length}
% %\vspace{-4mm}
% %\vs{5}
% \end{table}

\label{appendix:adding}

\subsection{Moving MNIST}
\label{appendix:moving_appendix}
We use the Stochastic Moving MNIST (SM-MNIST) \citep{denton2018stochastic} dataset which consists of sequences of frames of size $64\times 64$, containing one or two MNIST digits moving and bouncing off the walls. Training sequences are generated on the fly by sampling two different MNIST digits from the training set (60k total digits) and two distinct trajectories. Trajectories change randomly every time a digit hits a wall.

We trained the model on sequences only containing digits 0-7, and validate it on sequences with digits 8 and 9.  We ran the proposed algorithm with 2 layers, each with 6 RIMs and 4 active RIMs. We used a batch size of 100, and ran 600 batches per epoch (60000 samples per epoch). We trained for 300 epochs using Adam as optimizer \citep{Kingma2014} with a learning rate of 0.002.

\subsection{Bouncing Balls}
\label{appendix:bouncing_balls}
We use the bouncing-ball dataset from \citep{van2018relational}. The dataset consists of 50,000 training examples and 10,000 test examples showing $\sim$50 frames of either 4 solid balls bouncing in a confined square geometry, 6-8 balls bouncing in a confined geometry, or 3 balls bouncing in a confined geometry with a random occluded region.
In all cases, the balls bounce off the wall as well as off one another.  We train baselines as well as proposed model for about 100 epochs using 0.0007 as learning rate and using Adam as optimizer \citep{Kingma2014}. We use the same architecture for encoder as well as decoder as in \citep{van2018relational}. We train the proposed model as well as the baselines for 100 epochs. 

\begin{figure}[h]
  \centering
  \includegraphics[width=0.5\linewidth]{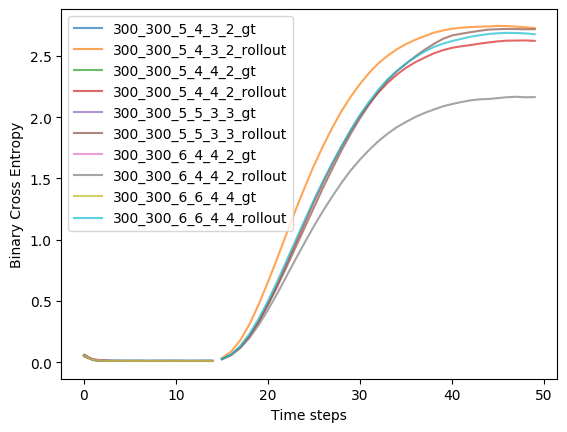}
  \caption{We show the performance of the proposed model for different values of number of RIMs on each layer, as well as number of active RIMs. The first 15 frames of ground truth are fed in and then the system is rolled out for the next 35 time steps.  The legend format is: (number of units on first layer, number of total units on second layer, total number of RIMs on first layer, total RIMs on second layer, activated RIMs on first layer, activated RIMs on second layer).  The best result is achieved by using sparse activation on both the higher and lower layer.  } 
  \label{fig:bouncing_balls_appendix}
\end{figure}

%\begin{figure}
%  \centering
%  \includegraphics[width=1.0\linewidth]{images/bouncing_balls_vis_0.png}
%  \caption{Vizualization of the proposed model for different values of number of RIMs on each layer, as well as number of active RIMs. We vary the number of RIMs in both layers, as well as the number of active RIMs. The first 5 frames of ground truth are fed in and then the system is
%rolled out for the next 15 time steps. Each row shows the visualization for the rollouts for 15 time steps. The descriptor at the beginning of each row gives: hidden units on the first layer, hidden units on the second layer, number of RIMs on the first layer, number of RIMs on the second layer, activated RIMs on the first layer, activated RIMs on the second layer.  } 
%  \label{fig:bouncing_balls_appendix}
%  \vspace{-6mm}
%\end{figure}

\subsection{Language Modelling}
Hyperparameters are  tuned for the vanilla LSTM, which consist of
{1, 2} LSTM layer, 0.3 embedding dropout, no layer norm, and shared, input/output embedding parameters. We use a hidden state of the size 2048
for the vanilla LSTM. Training was performed with Adam at learning rate 1e-3, gradients clipped to 0.1, sequence length 128, and batch size 128. 
We ran the proposed algorithm with 6 RIMs on each layer and kept the number of activated RIMs to 4 on each layer, with number of layers = 2. We have not done any hyper-parameter search for these experiments either for the baseline or for the proposed model.

\subsection{Atari}
\label{appendix:atari}
We used an open-source implementation of PPO from \citep{pytorchrl} with default parameters. We ran the proposed algorihtm with 6 RIMs on each layer and kept the number of activated RIMs to 4 on each layer, with number of layers = 2. We have not done any hyper-parameter search for Atari experiments.

\section{Pseudo-Code}

\begin{algorithm}[H]
\SetAlgoLined
\DontPrintSemicolon
\SetKwFunction{FMain}{BRIMsCell}
\SetKwFunction{FAtt}{Input Attention}
\SetKwFunction{FComm}{Communication}
\SetKwProg{Fn}{Function}{:}{}
\KwResult{RNN Cell forward for L layered BRIMs}
  \SetKwProg{Pn}{Function}{:}{}
    $x$ : Input \;
    $h_l$ : Hidden state of layer l represented as flat vector \;
    $h_l[k]$ : Hidden state of $k^{th}$ module of layer l \;
    $n_l$ : Number of modules in layer $l$\;
    $m_l$ : Number of modules kept active in layer $l$\;
    $\phi$ : Null vector \;
    All \texttt{Query}, \texttt{Key}, \texttt{Value} networks are fully connected neural networks\;
    $\mathbf{Q, K, V, A_S, A_R}$ denote matrices \;
    \textbf{Note:} Unless specified, all indexing start with 1\;
    \texttt{\\}
  \Pn{\FMain{\text{x}, $\text{h}_1$, ..., $\text{h}_{n_l}$}}{
    $h_0 = x$ \;
    \texttt{\\}
    \For{l = 1 to L} {
    \texttt{\\}
        \For{k = 1 \texttt{to} \tikzmark{top} $n_l$} {
            \textbf{Q}[k] = $\texttt{Input Query}_{l,k}(h_l[k])$ \;
        }
        \textbf{K}[0], \textbf{V}[0] = $\texttt{Null Key Value}_l(\phi)$ \hspace*{20em} \tikzmark{right}\; 
        \textbf{K}[1], \textbf{V}[1] = $\texttt{Input Key Value}_l(h_{l-1})$\;
        \textbf{K}[2], \textbf{V}[2] = $\texttt{Top-Down Key Value}_l(h_{l+1})$ \hspace{20pt}(if available) \;
        $\mathbf{A}_S$ = Softmax($\mathbf{QK}^T / \sqrt{d_{att}}$ ) \;
        $\mathbf{A}_R$ = \textbf{SV} \tikzmark{bottom}\;
        \AddNote{top}{bottom}{right}{Communication \\Between\\ Layers}
        \texttt{\\}
        % \For{k = 1 \texttt{to} \tikzmark{top} $n_l$} {
        %     Q[k] = $\textbf{Input Query}_{l,k}$( h[l, k]) \;
        % }
        % K[0], V[0] = $\textbf{Null Key Value}_l(\phi)$ \hspace*{20em} \tikzmark{right}\; 
        % \For{k = 1 \texttt{to} $n_{l-1}$}{
        %     K[k], V[k] = $\textbf{Input Key Value}_{l,k}$(h[l-1, k])\;
        % }
        % \For{k = 1 \texttt{to} $n_{l+1}$}{
        %     K[$n_{l-1} + k$], V[$n_{l-1} + k$] = $\textbf{Top-Down Key Value}_{l,k}$(h[l+1, k]) \hspace{20pt}(if available) \;
        % }
        % S = Softmax($QK^T / \sqrt{d_{att}}$ ) \;
        % input = SV \tikzmark{bottom}\;
        % \AddNote{top}{bottom}{right}{Communication \\Between\\ Layers}
        % \texttt{\\}
        \text{Sort $\mathbf{A}_S$[:,0] and take lowest $m_l$ as active} \tikzmark{top}\;
        \For{k s.t. module k is active}{
        $h_l[k]$ = $\texttt{RNN}_{l,k}$($\mathbf{A}_R[k], h_l[k])$ \hspace{53pt}(can use GRU or LSTM) \;
        }\tikzmark{bottom}
        \AddNote{top}{bottom}{right}{Sparse \\Activation}
        \texttt{\\}
        \For{k = 1 \texttt{to} \tikzmark{top} $n_l$}{
        \textbf{Q}[k] = $\texttt{Communication Query}_{l,k}(h_l[k])$ \;
        \textbf{K}[k] = $\texttt{Communication Key}_{l,k}(h_l[k])$ \;
        \textbf{V}[k] = $\texttt{Communication Value}_{l,k}(h_l[k])$ \;
        }
        $\mathbf{A}_R$ = Softmax($\mathbf{QK}^T / \sqrt{d_{att}}$) \,\textbf{V} \;
        \For{k s.t. module k is active}{
        $h_l[k]$ += $\mathbf{A}_R[k]$ \;
        }
        \tikzmark{bottom}
        \AddNote{top}{bottom}{right}{Communication \\Within\\ Layer}
    }
    \KwRet h \;
  }
 \caption{Single recurrent step for an L layered BRIMs model}
\end{algorithm}

% \subsection{Additional Figures}
% \begin{figure*}[]
% \vspace{0.2cm}
% \begin{minipage}{1.0\textwidth}
%     \centering
%     \includegraphics[width=1.0\linewidth]{figures/Communication.pdf}
%     \caption{}
% \end{minipage}
% \end{figure*}

% \begin{figure*}[]
% \vspace{0.2cm}
% \begin{minipage}{1.0\textwidth}
%     \centering
%     \includegraphics[width=1.0\linewidth]{figures/inp_att.pdf}
%     \caption{}
% \end{minipage}
% \end{figure*}

% \begin{align*}
%     \underbrace{\underbrace{\text{Softmax} \left(\begin{bmatrix}
%     \mathbf{Q}_{\;module\; 1} \\
%     \mathbf{Q}_{\;module\; 2} \\
%     \mathbf{Q}_{\;module\; 3} \\
%     \end{bmatrix}
%     \begin{bmatrix}
%     \mathbf{\color{myred}K}_{\;\phi} & \mathbf{\color{myblue}K}_{\;l-1} & \mathbf{\color{mygreen}K}_{\;l+1}
%     \end{bmatrix}
%     \right)}_{\bar{\mathbf{A}}^l_S}
%     \begin{bmatrix}
%     \mathbf{\color{myred}V}_{\;\phi} \\
%     \mathbf{\color{myblue}V}_{\;l-1} \\
%     \mathbf{\color{mygreen}V}_{\;l+1}
%     \end{bmatrix}}_{\bar{\mathbf{A}}^l_R} \\
%     \begin{matrix}
%     {\color{myred}\phi} &-& \text{Null}\\
%     {\color{myblue}l-1} &-& \text{Bottom Up}\\
%     {\color{mygreen}l+1} &-& \text{Top Down}\\
%     \end{matrix}
% \end{align*}
\end{document}